\documentclass[11pt]{article}

\usepackage[preprint]{acl}

\usepackage{newtxtext}
\usepackage{latexsym}
\usepackage[T1]{fontenc}
\usepackage[utf8]{inputenc}
\usepackage{microtype}
\usepackage{inconsolata}
\usepackage{graphicx}
\usepackage{amsmath}
\usepackage{amssymb}
\usepackage{booktabs}
\usepackage{multirow}
\usepackage{xspace}
\usepackage{xcolor}
\usepackage{float}
\usepackage{algorithm}
\usepackage{algpseudocode}

\newcommand{\email}[1]{{\fontfamily{zi4}\selectfont #1}}
\newcommand{\method}{\textsc{Edit}\xspace}
\newcommand{\methodA}{\textsc{Edit-SFT}\xspace}
\newcommand{\methodB}{\textsc{Edit-RL}\xspace}

\usepackage[most]{tcolorbox}  

\definecolor{prompthdr}{HTML}{222222}   
\definecolor{promptnote}{HTML}{8A8F98} 

\lstdefinestyle{promptstyle}{
basicstyle=\ttfamily\scriptsize, columns=fullflexible, keepspaces=true,
breaklines=true, breakindent=12pt, showstringspaces=false,
morecomment=[l][\color{prompthdr}\bfseries]{\#},   
morecomment=[s][\color{promptnote}\itshape]{[}{]}, 
}

\newtcblisting{promptbox}[1][]{
listing only, enhanced,
colback=white, colframe=black!45, boxrule=0.7pt,
arc=5pt, outer arc=5pt,
drop fuzzy shadow=black!30,
left=5pt, right=5pt, top=4pt, bottom=4pt,
listing options={style=promptstyle},
#1
}

\definecolor{cbad}{HTML}{C0392B}  
\definecolor{cgood}{HTML}{1E7E34}  
\definecolor{cbel}{HTML}{2E5597}

\newtcolorbox{casebox}[1][]{
enhanced, colback=white, colframe=black!55,
boxrule=0.6pt, arc=3pt, left=6pt, right=6pt, top=4pt, bottom=4pt,
fonttitle=\bfseries\small, coltitle=black, colbacktitle=black!8,
attach boxed title to top left={yshift=-2pt,xshift=6pt},
boxed title style={colframe=black!55,arc=2pt},
title={Case study: a belief-guided atomic rewrite (Private-Biology; content paraphrased)},
#1
}

\title{\textsc{Edit}: Evidence-Diagnosed Intervention Training \\for Rule-Faithful LLM Grading}

\author{
  \textbf{Zhihao Wu}$^{1}$\thanks{\ Equal contribution.} \quad
  \textbf{Linhai Zhang}$^{1}$\footnotemark[1] \quad 
  \textbf{Taiyi Wang}$^{2}$\footnotemark[1] \quad
  \textbf{Runcong Zhao}$^{1}$ \quad \\
  \textbf{Peter Andrews}$^{3}$ \quad 
  \textbf{Cesare Aloisi}$^{3}$ \quad
  \textbf{Yulan He}$^{1,4}$\thanks{\ Corresponding author.} \\[3pt]
  $^{1}$King's College London \quad
  $^{2}$University of Cambridge \quad
  $^{3}$AQA \quad
  $^{4}$The Alan Turing Institute \\[3pt]
  \email{\{zhihao.2.wu, linhai.zhang, runcong.zhao, yulan.he\}@kcl.ac.uk} \\
  \email{taiyi.wang@cl.cam.ac.uk}, \email{\{peter.andrews, caloisi\}@aqa.org.uk}
}

\begin{document}
\maketitle

\begin{abstract}
Reliable rubric grading requires more than accurate score prediction. Each judgement must be grounded in the mark scheme and evidence from the student answer. Existing credit-assignment and intervention methods, primarily designed for self-contained reasoning tasks such as mathematics reasoning, struggle in this setting because they do not identify where grading reasoning goes wrong or how the model's belief about the final mark changes during reasoning. We propose Evidence-Diagnosed Intervention Training (\method{}), a two-phase framework for training more rubric-faithful LLM graders. 
First, \methodA{} locates problematic reasoning steps using internal model signals: posterior belief over the final mark and input-grounding scores. It then revises only these local steps with help from a rubric checklist.
Second, \methodB{} calibrates the grader with belief-guided reward shaping, penalising large harmful belief drifts while still allowing helpful exploration.
Experiments on two real-world, multi-subject grading benchmarks demonstrate that \method{} consistently outperforms strong supervised fine-tuning and reinforcement learning baselines on both in-domain and out-of-domain splits, with ablation studies confirming that internal-state diagnostics drive these gains. Under deterministic rubric-edit interventions, \methodA{} is the most rule-responsive of all evaluated systems, and \methodB{} largely retains this responsiveness while improving accuracy.
\end{abstract}

\section{Introduction}
\label{sec:intro}

Grading a student's exam response is a rule application task. 
A grader needs to read the student answer, compare it with an explicit mark scheme, and assign a holistic mark. 
Large language models (LLMs) are increasingly deployed as automatic graders~\cite{grevisse2024llm,li-etal-2025-two, lai2025sasbench, wang2025large}. 
However, training them to grade reliably remains difficult. 
The final mark depends on a sequence of intermediate judgements, such as which rubric criteria are satisfied, what evidence supports them, which mark band applies, and how partial credits compose.

This makes rubric grading a demanding credit-assignment problem. 
Outcome-based reinforcement learning (RL) methods such as GRPO~\cite{shao2024deepseekmath} assign a single final reward to the whole reasoning trajectory~\cite{pignatelli2024a}, and they cannot identify which reasoning step caused an incorrect mark.
More fine-grained credit assignment methods, such as token-reallocation methods~\cite{jin2026dgpo} or intervention-style training~\cite{yang2026int}, offer a possible remedy. 

\begin{figure}[t]
    \centering
    \includegraphics[width=0.46\textwidth]{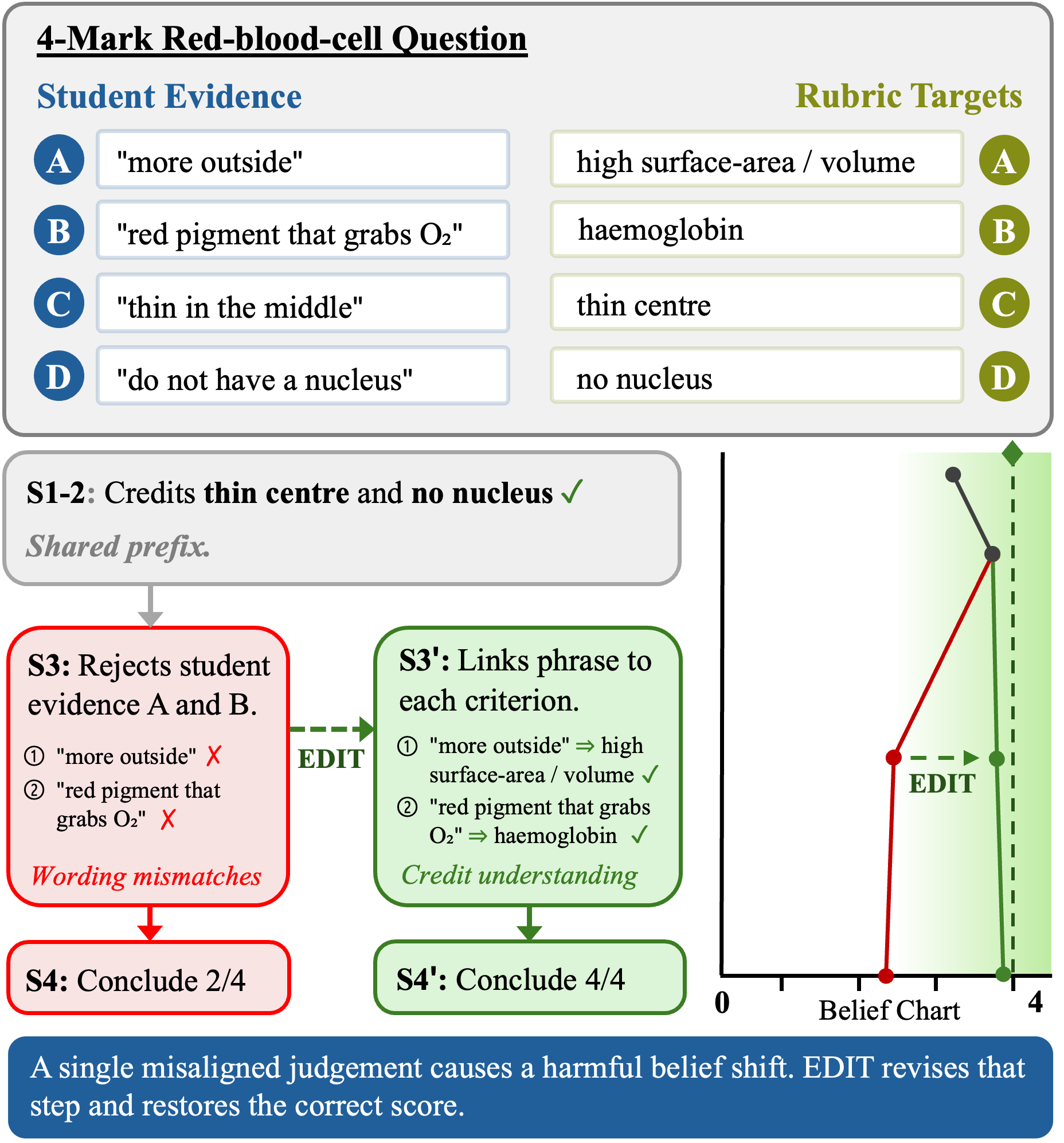}
    \caption{An illustration of a 4-mark grading question. The shared prefix (S1--2) correctly credits two criteria, but the standard grader then rejects two correct phrases due to wording mismatches (S3), shifting its belief away from the gold mark and concluding 2/4. \method{} locally revises only that step (S3$'$), linking each disputed phrase to its rubric criterion, which pulls the belief trajectory back to the gold mark and restores the correct mark.}
    \label{fig:intro}
\end{figure}

However, these methods were primarily designed for \textit{self-contained} reasoning tasks like mathematics, where the answer can often be checked by a verifier and the error is usually local. They are less suited to \emph{rule-application reasoning} tasks, where judging an answer requires checking whether each step correctly invokes and applies an external rule, rubric, or policy.
Rubric grading has two additional requirements. 
First, it requires \textbf{external grounding}. 
Rather than just maintaining internal logical consistency, each judgement step must be anchored in both the mark scheme and the student's written evidence.
Second, rule-based reasoning is an \textbf{uncertainty reduction} process. 
The model's belief about the final mark should progressively converge. Existing methods do not directly control this decision trajectory. Outcome-only RL ignores harmful belief shifts that occur in the middle of reasoning process, while token-level constraints can over-penalise benign exploratory phrasing.

To address these challenges, we introduce \method{} (Evidence-Diagnosed Intervention Training), a novel two-phase framework for training rubric-faithful LLM graders. 
As shown in Figure~\ref{fig:intro}, rather than relying on prompt-based self-audits or rigid token-level rewards, \method{} uses the model's internal decision signals to locate grading failures, revise them locally, and calibrate how the model's belief over the final mark evolves during reasoning.
Both phases read the same internal quantity, the model's posterior belief over the final mark. The first phase uses harmful transitions of this posterior to locate the step to repair, and the second constrains its trajectory during RL, so the supervision and the reward act in one shared coordinate system (\S\ref{sec:exp-main}).

To enforce strict external grounding, 
the first phase, \methodA{}, replaces prompt-based self-audits with \textbf{internal-state localisation}. At each reasoning step, it inspects the model's posterior belief over the final mark and computes mask-based grounding signals that measure whether the step depends on the student answer, the rubric, and the preceding reasoning steps. These signals identify where the reasoning first moves away from the correct mark or becomes weakly grounded. The selected step is then revised as a small atomic edit under a locality constraint. A rubric checklist is provided as privileged context, helping the reviser use the mark scheme without forcing the output to follow a rigid checklist format.

To manage the uncertainty reduction process, the second phase, \methodB{}, introduces \textbf{belief-guided reward shaping} for calibration RL. Instead of reallocating sequence-level reward across tokens, \methodB{} augments the standard mark-distance outcome reward with a penalty for large mid-trajectory belief drifts. Small fluctuations are allowed as benign exploration, but belief drifts that move too far from the gold mark are penalised in proportion to their excess.

Experimental results on two real-world, multi-subject student response grading benchmarks demonstrate that \method{} significantly outperforms strong SFT and RL baselines on both in-domain and out-of-domain splits. Furthermore, ablation studies confirm that our internal-state diagnostics are the primary source of these performance gains. 

In conclusion, our contributions are three-fold:
\begin{itemize}

\item We introduce \method{}, a novel two-phase \textbf{training framework for rubric-faithful grading}, that adapts credit assignment to externally grounded rubric grading rather than closed-system reasoning.

\item We replace unreliable prompt-based self-audits with robust posterior and grounding signals (\textbf{Internal State Diagnostics}), and use rubric-checklist context to support local atomic revisions without imposing a rigid output template.

\item We propose a calibration RL mechanism, called \textbf{Belief-Guided Reward Shaping}, that penalises severe mid-trajectory belief drifts while tolerating benign exploration, improving grading accuracy and out-of-domain generalisation.
\end{itemize}

\section{Methodology}
\label{sec:method}

\begin{figure*}[t]
\centering
\includegraphics[width=0.9\textwidth]{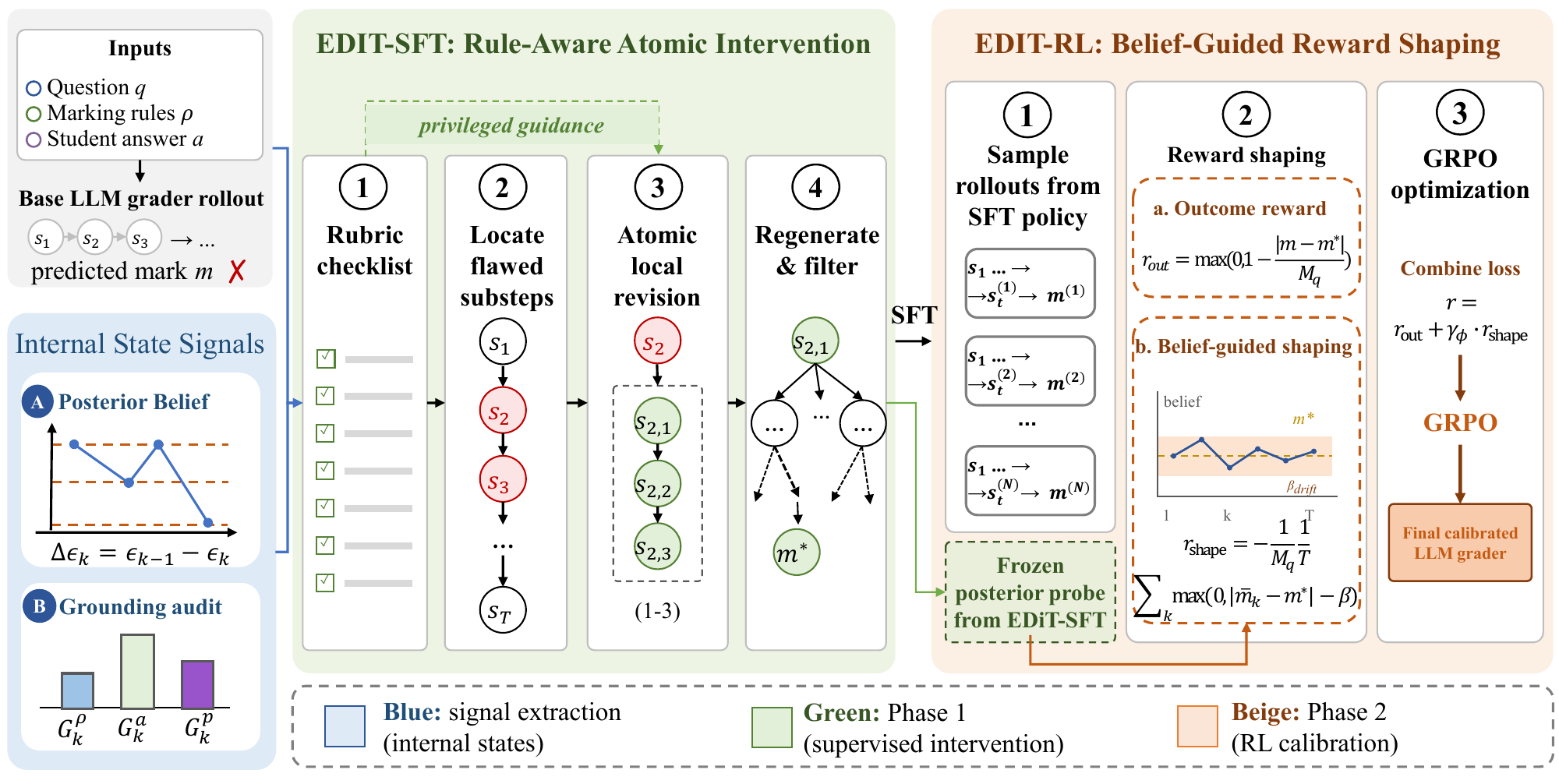}
\caption{The \method{} pipeline. \textbf{Phase~1 (\methodA{}):} For incorrect rollouts, internal-state signals, a posterior belief probe and masked-support grounding audits, locate and rank flawed substeps. A per-response rubric checklist is then provided to the reviser as privileged input (dashed arrow) to produce atomic corrective edits under a locality constraint, without imposing a rigid output template. These corrected trajectories are aggregated for supervised fine-tuning (SFT). \textbf{Phase~2 (\methodB{}):} The SFT policy is further calibrated using GRPO, augmented with a threshold-respecting, belief-guided reward (Eq.~\ref{eq:prs}). This shaping signal is computed from a frozen posterior probe to penalise mid-trajectory excursions.}
\label{fig:pipeline}
\end{figure*}

\method (Evidence-Diagnosed Intervention Training) fine-tunes an LLM grader in two phases (Fig.~\ref{fig:pipeline}). Phase~1 (\methodA, \S\ref{sec:method-phase-a}) builds a supervised training set by revising small, local parts of incorrect reasoning steps. The \emph{atomic} steps to revise are chosen using the model's own internal belief signals and grounding signals. The revised content is guided by a \emph{rubric-checklist}, which is extracted from the mark scheme and the student answer. Phase~2 (\methodB, \S\ref{sec:method-phase-b}) then calibrates the resulting policy with group-relative outcome-RL. The reward is shaped by the posterior belief signal used in Phase~1, giving a marking-task-aligned formulation that we call \emph{Belief-guided Reward Shaping}.

\subsection{Internal State Signals}
\label{sec:method-signals}

\method first extracts two step-level internal-state signals from the grading model (i.e., policy), both computed offline from existing rollouts. Consider a rollout be generated for the question $q$, marking rules $\rho$, and a student answer $a$. The rollout contains $T$ reasoning steps $s_{1:T}$. For each step boundary, we compute two signals.

\paragraph{Posterior belief probe.}
At step boundary $k\!\in\!\{0,\dots,T\}$, we estimate the policy's belief about the final integer mark, $m\!\in\!\{0,\dots,M_q\}$, where $M_q$ is the maximum available mark for question $q$. To do this, we append the final marking \emph{scaffold} to the partial reasoning chain and read the next-token logits over the valid mark tokens:
\begin{equation}
\label{eq:probe}
\hat p_k(m) \propto \exp(\mathrm{logit}_\theta(m | q, \rho, a, s_{1:k}, \textit{scaff.}))
\end{equation}
From $\hat p_k$, we derive the expected mark
$\hat m_k = \mathbb{E}_{m \sim \hat p_k}[m]$ and its absolute error $\epsilon_k
= |\hat m_k - m^*|$, where $m^*$ is the gold mark. We then define the signed per-step change as:
\begin{equation}
\label{eq:eegain}
\Delta \epsilon_k
= \epsilon_{k-1} - \epsilon_k
\end{equation}
A positive value means that step $s_k$ moved the policy's belief
\emph{closer} to the gold mark, while a negative value means that the step moved the belief \emph{farther away}. Thus, the sign shows whether the reasoning step locally helped or hurt the model's marking belief.

\paragraph{Mask-based grounding audit.}
For each reasoning step $s_k$, we compute mask-based support scores that
measure how much the likelihood of $s_k$ depends on three information sources:
the marking rule $\rho$, the student answer $a$, and the reasoning prefix
$s_{<k} = (s_1,\ldots,s_{k-1})$. Let $\ell_k(q,\rho,a,s_{<k})
= -\log p_\theta(s_k \mid q,\rho,a,s_{<k})$
denote the negative log-likelihood of step $s_k$ under the full context. We
write
\[
\ell_k^{\mathrm{full}}
= \ell_k(q,\rho,a,s_{<k}).
\]
The support scores are then defined as
\begin{align}
G_k^{\rho}
&= \ell_k(q,\texttt{[MASK]},a,s_{<k}) - \ell_k^{\mathrm{full}}, \notag\\
G_k^{a}
&= \ell_k(q,\rho,\texttt{[MASK]},s_{<k}) - \ell_k^{\mathrm{full}}, \notag\\
G_k^{p}
&= \ell_k(q,\rho,a,\varnothing) - \ell_k^{\mathrm{full}}.
\label{eq:support}
\end{align}

Larger values indicate that $s_k$ is more strongly supported by the corresponding masked input source. For example, if masking the student answer makes the step much harder to predict, then the step is likely grounded in the student answer. Because $G_k^p$ naturally increases for later steps (later steps tend to depend more on the earlier reasoning prefix), we use a position-residualised $\widetilde{G}_k^{p}$ as the reasoning-prefix-support score.
Let $u_{i,k}=k/T_i$ denote the relative position of step $k$ in example $i$, and
let
$\mu_p(u)=\mathbb{E}_{(i,j)\sim\mathcal{D}}
\!\left[
G_{i,j}^{p}
\,\middle|\,
u_{i,j}=u
\right]
$
be the dataset-level mean prefix-support score at relative position $u$.
We then define
\begin{equation}
\label{eq:pg_resid}
\widetilde{G}_{i,k}^{p}
=
G_{i,k}^{p}
-
\mu_p(u_{i,k}).
\end{equation}

This audit is an \emph{input-ablation} probe. We replace one input section with a placeholder and measure how much harder it becomes to predict the already-generated step. We validate this audit against external per-step labels produced by a stronger annotator model, as reported in Appendix~\ref{app:support-validation}. The grounding-audit scores correlate with these labels in the expected direction, and Appendix~\ref{app:counterfactual} further validates the causal utility of the selected steps.

\subsection{Phase 1: \methodA}
\label{sec:method-phase-a}

\methodA{} builds a supervised training pool by revising small parts of incorrect rollouts. Each rollout passes through four sub-stages: rubric-checklist generation, candidate step selection using internal signals, atomic revision, and outcome filtering. Successful revisions are then mixed with naturally correct rollouts and used as SFT targets.

\paragraph{Rubric-checklist.}
For each sample $x_i = \{q_i, \rho_i, a_i\}$, we generate a rubric-checklist by prompting the policy  with the gold mark $m^*_i$ as privileged information (available only during training-pool construction, not at inference). The policy is asked to list every marking point verbatim and to judge whether each point is covered, supported by a quoted evidence span. This checklist is later used as context during intervention.

\paragraph{Candidate steps selection.}
In this phase, we locate steps in an incorrect rollout ($m \neq m^*$) that should be edited, using the internal-state signals of \S\ref{sec:method-signals}. Steps are first \emph{shortlisted by the belief-shift criterion alone}: we rank steps by the most-negative $\Delta\epsilon_k$, discard positive or near-tie values, and keep the top-$k$ that most strongly moved the belief away from the gold mark. A step with a large negative belief shift but strong grounding, where the evidence was available yet the judgement still went wrong, is thus fully eligible and is our primary target. The grounding audit then \emph{re-ranks within this shortlist}, promoting steps whose $G_k^{\rho}, G_k^{a}$, and $\widetilde{G}_k^{p}$ all fall below the dataset-wide 25th percentile ahead of the others before the final candidate selection. The audit never excludes a step. Algorithm~\ref{alg:selection} (Appendix~\ref{app:selection}) gives the full procedure, Appendix~\ref{app:counterfactual} validates this rule against random and belief-shift-only alternatives, and implementation details are in Appendix~\ref{app:impl}.

\paragraph{Atomic revision under locality constraint.}
For each selected candidate step $s_k$, we ask the policy to revise it into 1--3 new sub-steps that fix only the local reasoning error. The prompt includes the sample context $\{q, \rho, a\}$, the full reasoning attempt with $s_k$ highlighted, the internal-signal diagnostic, and the previously generated rubric checklist. 

The revision must satisfy a locality constraint, ensuring that the intervention does not destroy the reasoning structure or coherence of the original rollouts. The constraint also prevents the revised step from simply copying dense information or oracle knowledge from the gold mark or rubric checklist. The details for the prompt and checklist-mimicry detector are shown in Appendix~\ref{sec:appendix_prompt}.

\paragraph{Pool composition and training.}
For each revision, we sample $N$ continuations and retain the revision if any continuation reaches the correct mark, $m = m^*$. The surviving rollouts are used as SFT samples. To preserve the policy's performance on easy or already-correct cases, we also mix the originally successful rollouts in the training pool. The policy is then trained on this combined dataset.

\subsection{Phase 2: \methodB}
\label{sec:method-phase-b}

Initialised from the \methodA{} policy, \methodB{} applies distance-aware GRPO \citep{shao2024deepseekmath} augmented by a belief-guided reward shaping term. By re-using the same posterior probe, this shaping term calibrates marking accuracy while penalising harmful drifts in intermediate beliefs.

\paragraph{Distance-aware outcome reward.}
We use a dense reward based on mark distance:
\begin{equation}
r_\text{out} = \max\Bigl(0,\, 1 - \tfrac{|m - m^*|}{M_q}\Bigr)
\end{equation}
with $r_\text{out}=0$ on parse failure. This reward gives higher scores to rollout whose predicted mark $m$ is closer to the gold mark $m^*$, providing distance-sensitive signal for RL calibration.

\paragraph{Belief-guided reward shaping.}
After \methodA, the model has learned to avoid reasoning steps that cause large harmful drifts in its posterior marking belief. In \methodB, we strengthen this behaviour by adding a per-rollout shaping signal that uses the posterior probe to check whether the model's belief drifted too much from the gold mark at any intermediate step:
\begin{equation}
r_\text{shape} = - \frac{1}{M_q}\frac{1}{T} \sum_{k=1}^{T}\max\bigl(0, |\hat{m}_k - m^*| - \beta_\text{drift}\bigr),
\label{eq:prs}
\end{equation}
where $\hat{m}_k$ is the expected posterior mark at step boundary $k$ (Eq.~\ref{eq:probe}). The threshold $\beta_\text{drift}$ is an \emph{exploration tolerance}, measured in raw marks, which accepts small belief fluctuations that may reflect useful exploration. However, when the belief moves beyond this tolerance, the rollout is penalised in proportion to the size of excess drift. The total reward is:
\begin{equation}
r = r_\text{out} + \gamma_\Phi \cdot r_\text{shape}
\end{equation}
This reward is used directly in the distance-aware GRPO objective. As a result, a rollout can be penalised relative to others in the same group if it contains a large belief deviation, even when its final mark is close to $m^*$.

The posterior belief probe is run on the \emph{frozen} \methodA{} checkpoint rather than the active policy. This maintains a stationary shaping signal, reduces computational overhead, and prevents reward hacking. We validate this choice by proving a small KL divergence between the trained policy with the initialised reference in Appendix~\ref{app:probe-kl}.
Although this relaxes the strict policy-invariance of classic PBRS \citep{ng1999pbrs}, it provides a stable, task-aligned constraint that significantly enhances out-of-distribution generalisation.

\section{Experiment}
\label{sec:experiments}

\subsection{Experiment Setup}
\label{sec:setup}

\paragraph{Datasets}
We evaluate policy grading on two complementary benchmarks, covering five datasets in total. The first benchmark is \textbf{SAS}, drawn from SAS-Bench \citep{lai2025sasbench}, a Chinese short-answer scoring benchmark. We use its three diverse subjects (History, Geography, and Physics). The second benchmark is \textbf{Private-Science}\footnote{A proprietary dataset shared with us under a data-sharing agreement. Licensing and privacy constraints prevent public release.}, a proprietary collection of student responses to GCSE-level science exam questions in two subjects (Biology and Physics), each marked by trained examiners against an official mark scheme. The two benchmarks are deliberately complementary. SAS comprises many questions with only a handful of responses each, while Private-Science contains fewer questions, but each is answered by hundreds of students. Full dataset statistics are given in Appendix~\ref{app:data}.

\paragraph{Evaluation protocol}
For Private-Science, we use three splits: a train set, an \textbf{in-distribution (ID)} test set of held-out responses to questions that \emph{appear} in training, and an \textbf{out-of-distribution (OOD)} test set of responses to questions that are \emph{never} seen in training. Separating ID from OOD tests whether a grader generalises to unseen questions rather than fitting question-specific surface patterns. SAS has no ID split. Each SAS question receives only two to five responses, so we evaluate the SAS only on the OOD questions, treating the SAS dataset as a subject-level fitting question.

Following standard practice in automated scoring \citep{li-etal-2025-two}, we use quadratic weighted kappa (QWK) as the main metric, macro-averaged over questions. We also report exact-match accuracy, within-1 accuracy, and mean absolute error (MAE).

\paragraph{Baselines}
All methods use \textbf{Qwen3-8B} as the policy model, adapted with LoRA. Training details are reported in Appendix~\ref{app:impl}. We keep the backbone fixed across methods so that performance differences mainly reflect the training objective rather than model capacity.
We compare our method after each phase (\methodA and \methodB) against four
baselines:\\
(1) \texttt{Base}, the off-the-shelf Qwen3-8B grader.\\
(2) \texttt{GRPO} \citep{shao2024deepseekmath}, RL on standard group-relative distance-aware outcome-reward.\\
(3) \texttt{DGPO} \citep{jin2026dgpo}, RL with  distribution-guided token-level advantage reallocation.\\
(4) \texttt{InT} \citep{yang2026int}, using SFT and
RL stages based on self-proposed interventions. 

\method and \texttt{InT} contain an SFT stage followed by an RL stage, whereas \texttt{GRPO} and \texttt{DGPO} apply RL directly to the base model.

\begin{table*}[t]
\centering
\footnotesize
\setlength{\tabcolsep}{5pt}
\begin{tabular}{llccccccc}
\toprule
\multirow{2}{*}{\textbf{Subject}} & \multirow{2}{*}{\textbf{Metric}}
 & \multirow{2}{*}{Base} & \multicolumn{2}{c}{\textit{RL-only}} & \multicolumn{2}{c}{InT} & \multicolumn{2}{c}{\textbf{\method (Ours)}} \\
\cmidrule(lr){4-5}\cmidrule(lr){6-7}\cmidrule(lr){8-9}
 & & & GRPO & DGPO & SFT & RL & \methodA & \methodB \\
\midrule
\multirow{4}{*}{History} & ACC\,$\uparrow$ & 0.419 & 0.471 & 0.462 & 0.426 & 0.461 & 0.454 & \textbf{0.477} \\
 & within-1\,$\uparrow$ & 0.512 & 0.597 & 0.600 & 0.566 & \textbf{0.625} & 0.569 & 0.623 \\
 & MAE\,$\downarrow$ & 2.612 & 1.883 & 1.865 & 1.852 & \textbf{1.569} & 1.952 & 1.615 \\
 & QWK\,$\uparrow$ & 0.783 & 0.873 & 0.876 & 0.871 & 0.889 & 0.863 & \textbf{0.899} \\
\midrule
\multirow{4}{*}{Geography} 
 & ACC\,$\uparrow$ & 0.414 & 0.400 & 0.367 & 0.367 & 0.333 & \textbf{0.433} & \textbf{0.433} \\
 & within-1\,$\uparrow$ & 0.552 & 0.533 & 0.600 & 0.600 & 0.533 & \textbf{0.633} & 0.600 \\
 & MAE\,$\downarrow$ & 1.483 & 1.533 & 1.533 & 1.533 & 1.667 & \textbf{1.267} & 1.300 \\
 & QWK\,$\uparrow$ & 0.706 & 0.676 & 0.685 & 0.745 & 0.675 & \textbf{0.779} & 0.756 \\
\midrule
\multirow{4}{*}{Physics} & ACC\,$\uparrow$ & 0.333 & 0.311 & 0.289 & 0.311 & 0.356 & \textbf{0.378} & \textbf{0.378} \\
 & within-1\,$\uparrow$ & 0.511 & 0.489 & 0.489 & 0.444 & 0.444 & \textbf{0.556} & 0.533 \\
 & MAE\,$\downarrow$ & 1.756 & 2.133 & 1.956 & 2.444 & 2.022 & \textbf{1.600} & 1.644 \\
 & QWK\,$\uparrow$ & 0.693 & 0.518 & 0.602 & 0.427 & 0.630 & 0.701 & \textbf{0.729} \\
\midrule
\multirow{4}{*}{\textbf{SAS-Avg}} & ACC\,$\uparrow$ & 0.389 & 0.394 & 0.372 & 0.368 & 0.383 & 0.421 & \textbf{0.429} \\
 & within-1\,$\uparrow$ & 0.525 & 0.540 & 0.563 & 0.535 & 0.534 & \textbf{0.586} & \textbf{0.586} \\
 & MAE\,$\downarrow$ & 1.950 & 1.850 & 1.784 & 1.943 & 1.753 & 1.606 & \textbf{1.520} \\
 & QWK\,$\uparrow$ & 0.727 & 0.689 & 0.721 & 0.681 & 0.731 & 0.781 & \textbf{0.794} \\
\bottomrule
\end{tabular}
\caption{Main results on \textbf{SAS}. All SAS questions are unseen during
training, so every split is out-of-distribution (no ID split). Metrics are over
the combined out-of-distribution evaluation set, with the best result per row in \textbf{bold}.}
\label{tab:main-sas}
\end{table*}

\subsection{Grading Performance}
\label{sec:exp-main}

\begin{table*}[t]
\centering
\footnotesize
\setlength{\tabcolsep}{5pt}
\begin{tabular}{lllccccccc}
\toprule
\multirow{2}{*}{\textbf{Subject}} & \multirow{2}{*}{\textbf{Split}} & \multirow{2}{*}{\textbf{Metric}}
 & \multirow{2}{*}{Base} & \multicolumn{2}{c}{\textit{RL-only}} & \multicolumn{2}{c}{InT} & \multicolumn{2}{c}{\textbf{\method (Ours)}} \\
\cmidrule(lr){5-6}\cmidrule(lr){7-8}\cmidrule(lr){9-10}
 & & & & GRPO & DGPO & SFT & RL & SFT & RL \\
\midrule
\multirow{8}{*}{Private-Biology}
 & \multirow{4}{*}{ID}  & ACC\,$\uparrow$       & 0.456 & 0.491 & 0.470 & 0.437 & 0.448 & 0.485 & \textbf{0.507} \\
 &                      & within-1\,$\uparrow$  & 0.793 & 0.809 & 0.801 & 0.812 & 0.828 & 0.821 & \textbf{0.831} \\
 &                      & MAE\,$\downarrow$     & 0.821 & 0.753 & 0.790 & 0.838 & 0.803 & 0.756 & \textbf{0.717} \\
 &                      & QWK\,$\uparrow$       & 0.701 & 0.727 & 0.704 & 0.630 & 0.648 & 0.738 & \textbf{0.754} \\
\cmidrule(l){2-10}
 & \multirow{4}{*}{OOD} & ACC\,$\uparrow$       & 0.460 & 0.478 & 0.478 & 0.422 & 0.425 & 0.475 & \textbf{0.489} \\
 &                      & within-1\,$\uparrow$  & 0.785 & 0.825 & 0.817 & 0.837 & 0.836 & 0.816 & \textbf{0.840} \\
 &                      & MAE\,$\downarrow$     & 0.786 & 0.718 & 0.725 & 0.777 & 0.772 & 0.735 & \textbf{0.687} \\
 &                      & QWK\,$\uparrow$       & 0.524 & 0.535 & 0.533 & 0.494 & 0.508 & 0.548 & \textbf{0.575} \\
\midrule
\multirow{8}{*}{Private-Physics}
 & \multirow{4}{*}{ID}  & ACC\,$\uparrow$       & 0.438 & 0.489 & 0.495 & 0.454 & \textbf{0.512} & 0.506 & \textbf{0.512} \\
 &                      & within-1\,$\uparrow$  & 0.762 & 0.780 & 0.795 & 0.804 & \textbf{0.822} & 0.782 & 0.815 \\
 &                      & MAE\,$\downarrow$     & 0.906 & 0.823 & 0.801 & 0.813 & \textbf{0.782} & 0.812 & 0.795 \\
 &                      & QWK\,$\uparrow$       & 0.477 & 0.480 & 0.535 & 0.454 & 0.551 & 0.527 & \textbf{0.555} \\
\cmidrule(l){2-10}
 & \multirow{4}{*}{OOD} & ACC\,$\uparrow$       & 0.429 & 0.431 & 0.435 & 0.374 & 0.433 & \textbf{0.452} & 0.445 \\
 &                      & within-1\,$\uparrow$  & 0.792 & 0.808 & 0.793 & 0.784 & 0.815 & 0.813 & \textbf{0.817} \\
 &                      & MAE\,$\downarrow$     & 0.829 & 0.805 & 0.820 & 0.950 & 0.840 & 0.784 & \textbf{0.772} \\
 &                      & QWK\,$\uparrow$       & 0.456 & 0.465 & 0.449 & 0.393 & 0.409 & 0.472 & \textbf{0.487} \\
\bottomrule
\end{tabular}
\caption{Main results on \textbf{Private-Science}. $\uparrow$/$\downarrow$ indicate higher/lower is better. Best fine-tuning result per row in \textbf{bold} (per-subject rows). }
\label{tab:main-aqa}
\end{table*}

We first discuss the public SAS benchmark, where every evaluation question is unseen during training. As shown in Table~\ref{tab:main-sas}, \methodB{} achieves the best results on average across all metrics. In particular, it improves macro-QWK from $0.727$ for \texttt{Base} and $0.731$ for \texttt{InT} to $0.794$. These results show that the method improves grading accuracy in a fully OOD setting.

The subject-level results show some variation. \methodB{} gives the best QWK on History and Physics. \texttt{InT-RL} remains strongest on History for several metrics, while \methodA{} performs best on Geography, where all RL variants degrade the performance. However, \methodB remains the only RL calibrated model that strongly outperforms \texttt{Base}. Appendix~\ref{app:variance} traces most of this variation to final score-band and credit-synthesis errors. For example, \texttt{InT-SFT} never withholds full credit on zero-score Physics responses, whereas \methodA{} correctly does so on 64\% of them.
Nevertheless, our full method provides the most consistent overall improvement across SAS.

Because the smallest SAS subjects contain only 25--30 test responses, we further assess significance at the question level, pooling all 37 held-out questions with equal weight. Intervals use a cluster bootstrap over questions, comparisons against \texttt{Base} are paired, and Holm correction is applied across the six comparisons (Appendix~\ref{app:significance}). As shown in Table~\ref{tab:sas-sig}, baseline gains near $0.03$ do not survive correction, whereas \method{}'s gains are two to three times larger and remain significant (both $p{=}0.0006$).

\begin{table}[t]
\centering
\footnotesize
\setlength{\tabcolsep}{4pt}
\begin{tabular}{lcc}
\toprule
\textbf{Method} & \textbf{QWK} [95\% CI] & \textbf{$\Delta$ vs.\ Base} (Holm $p$) \\
\midrule
\texttt{Base}    & 0.760 [0.677, 0.830] & -- \\
\texttt{GRPO}    & 0.794 [0.707, 0.867] & $+0.034$ ($p{=}0.391$) \\
\texttt{DGPO}    & 0.809 [0.732, 0.872] & $+0.049$ ($p{=}0.021$) \\
\texttt{InT-SFT} & 0.793 [0.707, 0.868] & $+0.033$ ($p{=}0.391$) \\
\texttt{InT-RL}  & 0.822 [0.750, 0.883] & $+0.062$ ($p{=}0.051$) \\
\methodA         & 0.831 [0.764, 0.887] & $+0.072$ ($p{=}\textbf{0.0006}$) \\
\methodB         & \textbf{0.856} [0.784, 0.913] & $+0.096$ ($p{=}\textbf{0.0006}$) \\
\bottomrule
\end{tabular}
\caption{Question-level pooled QWK on SAS with cluster-bootstrap 95\% CIs and Holm-corrected $p$-values against \texttt{Base} (Appendix~\ref{app:significance}).}
\label{tab:sas-sig}
\end{table}

We next evaluate Private-Science, where both ID and OOD splits are available. 
As shown in Table~\ref{tab:main-aqa}, \methodA{} achieves the best QWK among all baselines on Private-Biology, improving macro-QWK by $+0.011$ on the ID-test split and $+0.013$ on the OOD-pool over the strongest baseline (\texttt{GRPO}). \texttt{InT-RL} outperforms across multiple metrics in Private-Physics ID setting, while degrading drastically in the OOD, suggesting an overfitting on the trained types of questions. In contrast, \methodA shows steady gains across both ID and OOD. This suggests that the improvement is not simply due to fitting question-specific patterns seen during training. 
We attribute the improvement from \methodA{} to three design choices. First, internal signals help locate reasoning steps that are likely to need revision. Second, atomic revisions keep the intervention local and avoid disrupting the full reasoning chain. Third, the rubric-checklist context gives the revisor task-specific guidance without forcing it to copy the checklist directly. These designs together preserve the model's original reasoning capabilities while making its marking reasoning chain more grounded and better aligned with the rubric.

\methodB{} further improves performance over \methodA{} on both test pools. It improves macro-QWK by $+0.016$ and $+0.027$ on Biology's ID and OOD, respectively. It also improves ACC, within-1, and MAE, giving the best results across all metrics. \methodB{} attains the highest OOD QWK on both subjects and the best value in seven of eight OOD subject-by-metric cells. This suggests that belief-guided reward shaping is especially useful when the model must grade unseen questions.

\begin{figure}[t]
\centering
\includegraphics[width=\columnwidth]{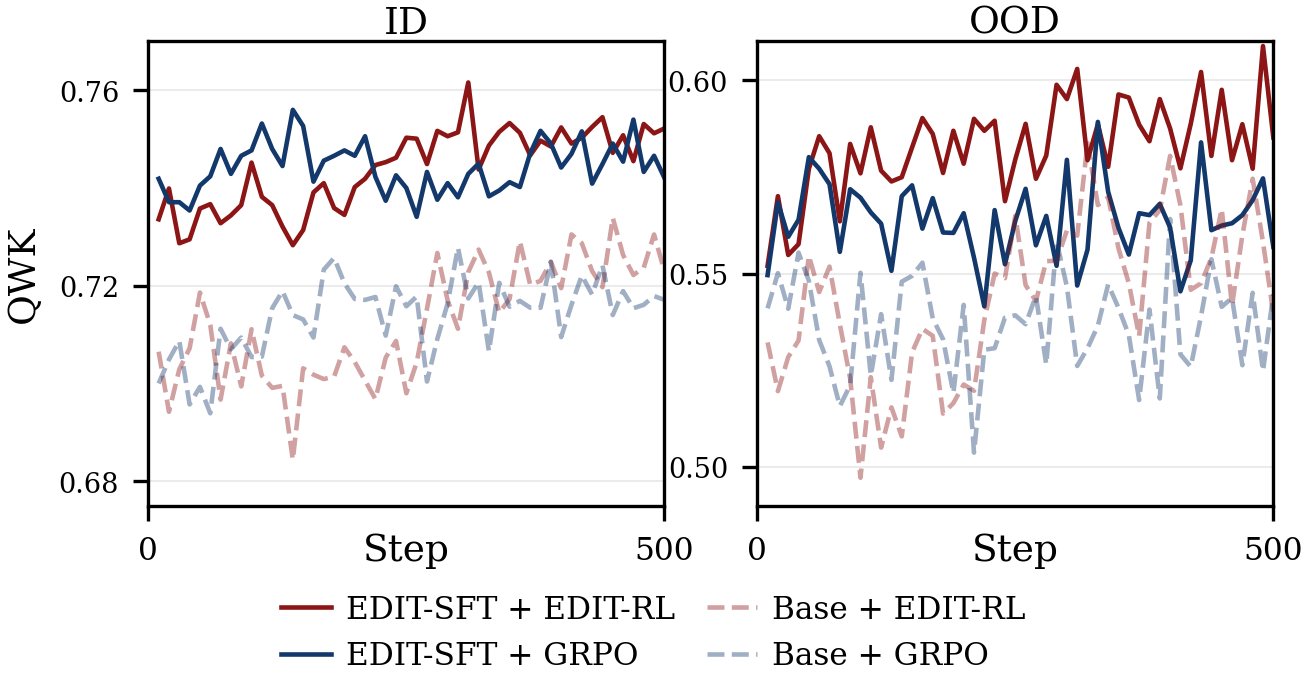}
\caption{Training dynamics of RL choices (GRPO, \methodB) with base choices (\texttt{Base}, \methodA) in QWK on ID and OOD test set. Curves are computed post hoc from saved checkpoints (Appendix~\ref{app:impl}).}
\label{fig:rl_comparison}
\end{figure}

Table~\ref{tab:sft_rl_comparison} further separates the effect of the SFT and the RL. Applying \methodB{} directly to \texttt{Base} does not improve OOD performance over standard GRPO. In contrast, applying RL after \methodA{} gives stronger results, and the full combination \methodA{}$+$\methodB{} is best on every metric. Figure~\ref{fig:rl_comparison} illustrates the QWK dynamics on the ID and OOD with RL proceeds. Although the improvement matches with a GRPO on \methodA in ID, \methodB shows steady improvement on the OOD setting. This suggests that belief-guided reward shaping is most effective when the model has first been trained on internally located, rubric-aware atomic revisions.

To isolate the effect of the belief-shaped reward from seed noise, Table~\ref{tab:rl-seeds} repeats this comparison over three seeds on two datasets. Both arms share the same \methodA{} checkpoint and identical budgets, decoding, and final checkpoint, differing only in the reward term. \methodB{} outperforms GRPO in all three paired runs on both datasets, and the paired 95\% CI excludes zero on Private-Biology.

\begin{table}[t]
\centering
\footnotesize
\setlength{\tabcolsep}{4pt}
\begin{tabular}{lccc}
\toprule
\textbf{Dataset} & \textbf{$+$GRPO} & \textbf{$+$\methodB} & \textbf{$\Delta$} \\
\midrule
Private-Biology & 0.559\,{\scriptsize$\pm$0.003} & \textbf{0.580}\,{\scriptsize$\pm$0.004} & $+0.021^{*}$ \\
SAS-History    & 0.873\,{\scriptsize$\pm$0.008} & \textbf{0.885}\,{\scriptsize$\pm$0.004} & $+0.012$ \\
\bottomrule
\end{tabular}
\caption{OOD QWK (mean\,$\pm$\,SD over three seeds). $\Delta$ is the mean seed-wise paired difference. $^{*}$ marks a paired 95\% CI excluding zero (CIs: $[+0.011, +0.030]$ and $[-0.007, +0.030]$).}
\label{tab:rl-seeds}
\end{table}

\begin{table}[h]
\centering
\footnotesize
\setlength{\tabcolsep}{4pt}
\resizebox{\columnwidth}{!}{%
\begin{tabular}{lcccc}
\toprule
\textbf{Method} & \multicolumn{2}{c}{\textbf{ID}} & \multicolumn{2}{c}{\textbf{OOD}} \\
\cmidrule(lr){2-3}\cmidrule(lr){4-5}
 & QWK\,$\uparrow$ & ACC\,$\uparrow$ & QWK\,$\uparrow$ & ACC\,$\uparrow$ \\
\midrule
\texttt{Base}              & 0.701 & 0.456 & 0.524 & 0.460 \\
\texttt{Base}~+~GRPO       & 0.727 & 0.491 & 0.535 & 0.478 \\
\texttt{Base}~+~\methodB   & 0.727 & 0.483 & 0.518 & 0.461 \\
\methodA~+~GRPO            & 0.748 & 0.506 & 0.560 & 0.484 \\
\methodA~+~\methodB        & \textbf{0.754} & \textbf{0.507} & \textbf{0.575} & \textbf{0.489} \\
\bottomrule
\end{tabular}%
}
\caption{RL choices (single seed) applied either directly on \texttt{Base} or on the
\methodA{} checkpoint, Private-Biology. The combination of \methodA{} SFT
and \methodB{} (belief-guided shaped GRPO) is best on every
metric/split.}
\label{tab:sft_rl_comparison}
\end{table}

\subsection{Ablation for Rule-Aware Atomic Intervention}
\label{sec:exp-ablation-locator}

We ablate the main design choices of \methodA{} on SAS-History, the largest dataset in the SAS suite we adopted. The ablation focuses on three components: internal-signal localisation, rubric-checklist context, and atomic locality constraints. These components test whether performance gains come from selecting better edit locations, giving the revisor better rule context, and keeping the revision local.

\begin{figure}[t]
\centering
\includegraphics[width=0.65\columnwidth]{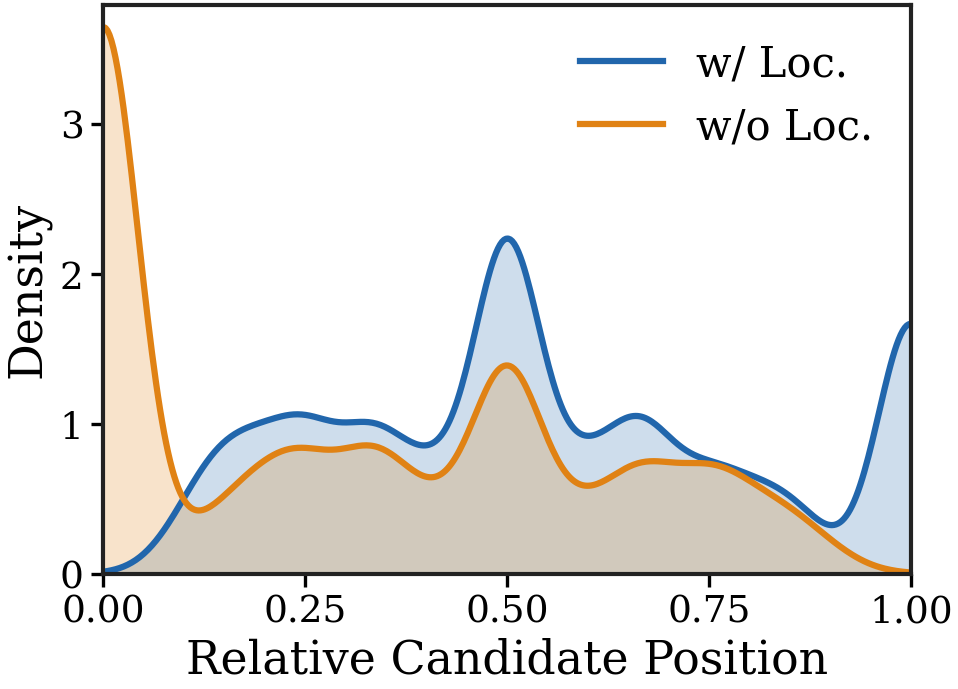}
\caption{Comparison between the relative positions of the candidate step selected with or without the internal-state driven localiser from \methodA.}
\label{fig:ablation-locator}
\end{figure}

As shown in Table~\ref{tab:ablation-sft}, removing any module degrades performance. Excluding the rubric-checklist context leads to the largest drop in QWK and MAE while leaving ACC largely unchanged, suggesting increased variance in grading behaviour. Removing atomic locality constraints during revision results in a slight relative degradation. This is likely because SAS scoring is naturally point-based and therefore already imposes contextual constraints. \methodA without internal-signal localisation exhibits the largest drop across all metrics, highlighting the critical role of our step-candidate selection for revision. Appendix~\ref{app:counterfactual} further shows counterfactually that \method{}-selected steps recover the gold mark more often than position-matched random or belief-shift-only selections.
Figure~\ref{fig:ablation-locator} further shows that the base model tends to select earlier steps, whereas our localisation module selects  middle or final steps, where reasoning synthesis typically occurs.

\begin{table}[h]
\centering
\small
\setlength{\tabcolsep}{5pt}
\begin{tabular}{lcccc}
\toprule
\textbf{Ablation} & QWK\,$\uparrow$ & ACC\,$\uparrow$ & Within-1\,$\uparrow$ & MAE\,$\downarrow$ \\
\midrule
\methodA       & \textbf{0.863} & \textbf{0.454} & \textbf{0.569} & \textbf{1.952} \\
w/o Localiser      & 0.820 & 0.408 & 0.515 & 2.300 \\
w/o Checklist      & 0.817 & 0.454 & 0.554 & 2.346 \\
w/o Atomic         & 0.842 & 0.446 & 0.562 & 2.046 \\
\bottomrule
\end{tabular}
\caption{Ablation of three core designs of \methodA on SAS-History, reported in all four metrics.}
\label{tab:ablation-sft}
\end{table}

\subsection{Rule-Faithfulness under Interventions}
\label{sec:exp-intervention}

We also test whether a grader \emph{truly follows the rubric}, rather than relying on its internalised scoring scale. To do this, we apply deterministic edits to the mark scheme where the gold-score change can be computed exactly. We then measure the Rule-Sensitivity Ratio (RSR), defined as the ratio between the
model's score change and the gold score change. ($\mathrm{RSR}{=}1$ means that the model follows the rule edit exactly; ${<}1$ under-reacts, ${>}1$ over-reacts). The edits include level-of-response band shifts (LoR) and points-total rescaling (PTS). The detailed intervention strategies are reported in Appendix~\ref{app:int-strategies}. One complication is that a predicted score of $0$ is unchanged by every edit, so we decompose RSR into two parts: 1. Floor Rate (FR), the fraction of predictions equal to zero, and 2. In-Rubric RSR (IB-RSR), responsiveness only among non-zero predictions. A further complication is that \emph{signed} IB-RSR means can hide errors, because over-reaction and under-reaction cancel when averaged across conditions. We therefore report AE-RSR, the mean of $|\mathrm{IB\text{-}RSR}-1|$ across edit conditions (lower is better).

Table~\ref{tab:int-decomp} averages results across the LoR and PTS intervention strategies. Under LoR band shifts, every grader is near-faithful. \methodA{} attains the lowest AE-RSR ($0.070$) and \methodB{} the lowest floor rate ($6.60\%$). The separation appears under PTS rescaling, where \texttt{Base} is the weakest system (AE-RSR $0.733$) and \methodA{} the strongest ($0.500$). Aggregated over all six conditions, \methodA{} reaches $0.285$ against $0.412$ for \texttt{Base}. \texttt{InT} attains PTS responsiveness comparable to \method{}'s, but partly by collapsing predictions to the floor ($28.57\%$ vs.\ $19.90\%$ for the SFT variants), which removes responsive rows from the in-rubric statistic.

Overall, \methodA{} is the most rule-responsive system, and \methodB{} trades a small amount of responsiveness for grading accuracy. Meanwhile, PTS rescaling remains unsolved for every evaluated system (AE-RSR ${\ge}0.5$), pointing to absolute-count anchoring as an open challenge for rubric grading. Our intervention protocol exposes exactly this kind of failure, which accuracy metrics alone cannot detect: a grader can score well while remaining insensitive to the rule it is meant to apply.

\begin{table}[t]
\centering
\footnotesize
\setlength{\tabcolsep}{3.5pt}
\begin{tabular}{lccccc}
\toprule
\textbf{Method} & \multicolumn{2}{c}{\textbf{LoR}} & \multicolumn{2}{c}{\textbf{PTS}} & \textbf{Overall} \\
\cmidrule(lr){2-3}\cmidrule(lr){4-5}\cmidrule(lr){6-6}
 & AE\,$\downarrow$ & FR\,(\%) & AE\,$\downarrow$ & FR\,(\%) & AE\,$\downarrow$ \\
\midrule
\texttt{Base}     & 0.090 & 10.07 & 0.733 & 22.43 & 0.412 \\
\texttt{GRPO}     & 0.093 & 7.17  & 0.710 & 18.80 & 0.402 \\
\texttt{DGPO}     & 0.090 & 7.20  & 0.697 & 19.33 & 0.393 \\
\texttt{InT-SFT}  & 0.093 & 10.57 & 0.530 & 28.57 & 0.312 \\
\texttt{InT-RL}   & 0.143 & 9.70  & 0.507 & 26.80 & 0.325 \\
\methodA          & \textbf{0.070} & 7.00 & \textbf{0.500} & 19.90 & \textbf{0.285} \\
\methodB          & 0.083 & 6.60 & 0.510 & 19.87 & 0.297 \\
\bottomrule
\end{tabular}
\caption{Rule faithfulness on Private-Biology. AE = AE-RSR (lower is better). FR = floor rate (\%). Overall averages AE-RSR over all six conditions, with per-condition signed results in Appendix~\ref{app:int_full_results}.}
\label{tab:int-decomp}
\end{table}

\section{Related Work}
\label{sec:related}

\paragraph{Credit Assignment for Reinforcement Learning}
Credit assignment refers to the problem of attributing a delayed outcome to specific intermediate decisions, which remains a core challenge in LLM reasoning.
Standard sequence-level objectives like GRPO~\citep{shao2024deepseekmath} broadcast a single scalar reward uniformly across a trajectory.
To refine this coarse signal, recent work generally falls into three paradigms.
Return redistribution methods, such as DGPO~\citep{jin2026dgpo}, reallocate trajectory-level returns to individual steps.
Process reward models (PRMs), like Math-Shepherd~\citep{wang2024mathshepherd}, train explicit step-level verifiers to provide fine-grained supervision.
PRMs derive the process signal from external step-level labels, whereas \method{} derives it from the model's own decision state under final-mark supervision only. Such labels are also rubric-specific, defined only relative to one mark scheme, so we regard PRM-style supervision as a complementary direction that requires per-rubric annotation.
Alternatively, hindsight interventions, such as InT~\citep{yang2026int}, prompt the policy to self-audit and rewrite erroneous steps.
However, a growing body of work shows that LLM-generated chains-of-thought frequently diverge from actual computation, often necessitating external rewards to enforce faithfulness~\citep{turpin2023unfaithful, lanham2023measuring}.
While prior methods rely on external verifiers or textual self-critique, our framework directly localises critical steps using the model's well-calibrated internal signals and guides the policy with belief-guided reward shaping.

\paragraph{LLM-based Automated Student Assessment}
Automated essay scoring (AES) has increasingly transitioned from training dedicated classifiers to utilizing pretrained LLMs as evaluators or distilling their rationales into smaller models~\citep{lingvincent2024aes, li-etal-2023-distilling}.
For short-answer and science grading, recent benchmarks and systems primarily rely on prompt engineering or verbal reflection of frozen models~\citep{lai2025sasbench, li-etal-2025-two}.
Other approaches explore aligning evaluation rationales via preference optimization on thought trees to improve explainability~\citep{li-etal-2024-calibrating}.
While most existing systems treat grading as a zero-shot prompting task or rely on external solvers to follow rules, our work frames grading as an end-to-end trainable credit-assignment problem.
By supplying the rubric checklist as a privileged input during the rewriting phase, our approach decouples analytical reasoning from rigid structural formats to ensure faithful adherence to external grading criteria.

\section{Conclusion}
\label{sec:conclusion}

In this paper, we propose \method{}, a novel two-phase framework addressing the unique credit-assignment challenges of rule-faithful LLM grading.
To enforce strict external grounding, \method{} replaces unreliable prompt-based self-audits with internal-state localisation and privileged-context rewriting.
Furthermore, our belief-guided reward shaping guides the uncertainty reduction process by penalising severe belief excursions while tolerating benign exploration.
Experiments on two real-world grading datasets demonstrate that \method{} significantly outperforms state-of-the-art baselines across in-domain and out-of-domain splits.
Future work will explore generalising the framework to broader domains, streamlining the training pipeline, and dynamically decomposing holistic rubrics.


\section*{Limitations}

We identify four key limitations of the \method{} framework.

First, our evaluation scenarios are currently restricted in scope. Although we frame student answer scoring as a complex rule-based reasoning task, we evaluate \method{} on two datasets primarily consisting of short-answer and fill-in-the-blank questions. We have not yet tested our framework on long-form essay scoring tasks, such as those in the Automated Student Assessment Prize (ASAP) benchmark, nor have we explored its applicability in other high-stakes rule-based reasoning domains like legal judgment or medical diagnosis. In addition, the Private-Science OOD splits contain only two to three questions per subject, so our question-level significance analysis is limited to SAS. An important direction for future work is to verify and extend the generalisability of our method across broader domains and longer, more unstructured text formats.

Second, the proposed training pipeline involves considerable complexity. While our probing-based internal-state localisation reduces inference overhead compared to prior prompt-elicited self-audits, the overall framework still relies on a heavy multi-stage process: rigorous SFT data construction followed by a calibration RL phase (GRPO). This multi-step orchestration increases both computational complexity and engineering effort. In addition, while the two RL arms in our controlled comparison are budget-matched, the pipeline-level comparison against baselines is not: \method{} incurs extra offline compute for posterior-belief probing, grounding audits, and revision sampling (Appendix~\ref{app:impl}). Future research should investigate more streamlined, end-to-end training paradigms that simplify the pipeline while preserving the precise credit-assignment capabilities of evidence-diagnosed interventions.

Third, \method{} relies on the availability and granularity of explicit rule sets. Specifically, \methodA{} utilises a per-criterion rubric checklist as privileged input to constrain atomic rewrites. However, in many real-world settings, rubrics can be vague, holistic, or implicitly defined by human graders. The effectiveness of our internal-state diagnostics and rule-faithful rewriting may degrade when applied to such unstructured or noisy criteria. Future work could explore mechanisms to dynamically decompose holistic rubrics into structured checklists, or adapt the framework to handle fuzzy and implicit rule constraints.

Finally, like other outcome-filtered self-improvement methods in the STaR and ReST family \citep{zelikman2022star,gulcehre2023rest}, \methodA{} retains only revisions whose continuations reach the gold mark: 62.8\% of incorrect rollouts on Private-Biology and 76.4\% on SAS-History yield no corrected example (Appendix~\ref{app:counterfactual}). This costs sampling compute rather than data validity, since every retained trajectory is verified against the gold mark, but the outcome filter does bias the retained pool toward cases that are repairable by a local edit.

\section*{Ethics Statement}
This research has been approved by our institutional ethics committee. Our dataset comprises LLM-augmented open-source question banks and proprietary data from a collaborative institution (which has also passed their internal rigorous ethical review). All data has been strictly de-identified to remove Personally Identifiable Information (PII). Given that the open-source subset relies on LLM generation, it may inherit sociopolitical or linguistic biases from pre-training corpora. We strongly recommend manual review before using this data for downstream training to avoid penalising diverse linguistic expressions or propagating cognitive distortions.

Furthermore, our framework is intended solely for researching LLM reasoning in rule-constrained environments. Automated grading involves high-stakes decisions with profound psychological impacts on students' academic trajectories. We strictly advise against deploying these models in real-world educational settings without human-in-the-loop to ensure fairness, accountability, and pedagogical empathy.

\section*{Acknowledgements}
This work was supported in part by the UK Engineering and Physical Sciences Research Council (EPSRC) through the Prosperity Partnership scheme (grant no. UKRI566) and a Turing AI Fellowship (grant no. EP/V020579/1, EP/V020579/2).

\bibliography{custom}

\appendix
\section{Dataset Statistics}
\label{app:data}

Table~\ref{tab:data} reports, for each dataset, the number of responses and
distinct questions in every split, together with the range of per-question
maximum marks. The three splits play different roles across the two benchmarks.
For \textbf{Private-Science}, each question is answered by hundreds of students: the
\emph{Test} split holds out responses to questions that are \emph{seen} during
training (in-distribution, ID), whereas the \emph{Holdout} split consists of
entirely \emph{unseen} questions (out-of-distribution, OOD). For \textbf{SAS},
each question has only a handful of responses, so questions are partitioned
disjointly across all three splits; \emph{both} the Test and Holdout splits are
therefore out-of-distribution, and SAS has no ID split.

\paragraph{Data availability.}
The Private-Science responses were fully anonymised by the data provider before being shared with us. We report experimental results on this dataset, but it cannot be publicly released owing to licensing and privacy constraints. The SAS benchmark is publicly available \citep{lai2025sasbench}.

\begin{table*}[t]
\centering
\small
\begin{tabular}{llrrrc}
\toprule
Benchmark & Dataset & Train (resp / q) & ID (resp / q) & OOD (resp / q) & Marks \\
\midrule
\multirow{3}{*}{Private-Science (proprietary)}
 & Biology     & 10{,}159 / 12 & 1{,}264 / 12 & 2{,}087 / 2 & 3--6 \\
 & Physics     &  1{,}573 / 7  &   395 / 7    & 655 / 3 & 3--6 \\
\midrule
\multirow{5}{*}{SAS \citep{lai2025sasbench}}
 & History     & 510 / 102  & -- & 130 / 26 & 11--26 \\
 & Geography   &  110 / 22  & -- &  30 / 6  & 10 \\
 & Physics     &  95 / 19   & -- &  25 / 5  & 5--20 \\
\bottomrule
\end{tabular}
\caption{Dataset statistics. ``resp / q'' is the number of responses and the number of distinct questions in each split. For Private-Science, the ID split is in-distribution (held-out student responses to seen questions), and the OOD split is
out-of-distribution (unseen questions). For SAS, there is no ID split.}
\label{tab:data}
\end{table*}

\section{Implementation Details}
\label{app:impl}

\paragraph{Training}
All methods adapt Qwen3-8B using a shared LoRA recipe: rank $r{=}64$, $\alpha{=}128$, applied to all attention and MLP projection matrices ($q,k,v,o,\text{gate},\text{up},\text{down}$). We train for $2$ epochs at learning rate $2\times10^{-4}$ (cosine schedule, $10\%$
warmup). For the SAS dataset, we set the effective batch size to $8$ due to the small amount of data, and to $32$ on Private-Science. The same recipe is used for all subjects and methods, so comparisons are not confounded by hyperparameter differences.

\paragraph{Decoding and evaluation}
At evaluation, we obtain $1$ completion per marking using greedy sampling. This ensures consistency across all datasets and methods. QWK is computed per question over the integer score range $[0, M_q]$, where $M_q$ is the question's maximum mark, using quadratic weights, and then macro-averaged across questions.

\paragraph{RL configuration and checkpoint selection}
No test metric or test data enters hyperparameter or checkpoint selection at any point, and every reported result uses the final-step checkpoint rather than a selected peak. Training horizons are fixed in advance at $200$ RL steps for SAS and $500$ for Private-Science. All RL methods share one optimisation recipe (Table~\ref{tab:rl-hparams}): learning rate $5\times10^{-6}$ with a cosine schedule and $3\%$ warmup, KL coefficient $\beta_{\mathrm{KL}}{=}0.02$ towards the initialisation, $8$ rollouts per prompt and $4$ prompts per optimiser step ($32$ completions), sampling temperature $1.0$, and a completion budget of $1{,}024$ tokens. A fresh LoRA adapter with the SFT configuration ($r{=}64$, $\alpha{=}128$, all projection matrices) is trained on top of the merged initialisation, which also serves as the KL reference. The test-set curves in Figure~\ref{fig:rl_comparison} were generated post hoc after training had finished and were never used for monitoring or selection. The three-seed comparison in Table~\ref{tab:rl-seeds} repeats the full training run with seeds $42$, $43$, and $44$.

\methodB{} introduces two additional hyperparameters, the drift tolerance $\beta_\text{drift}$ and the shaping weight $\gamma_\Phi$. Both are calibrated per dataset on a random $10\%$ subset of training rollouts using signal statistics only, never a performance metric. We sweep $\beta_\text{drift}$ and keep the value that maximises the separation between the shaping-penalty firing rates on incorrect and correct rollouts, and we set $\gamma_\Phi$ to the largest value that remains \emph{flip-safe}: the accumulated penalty on a correct rollout stays below the outcome-reward gap to an incorrect one, so the shaping term can never invert the outcome ranking within a group. The posterior probe is read at temperature $0.7$. The selected values are listed in Table~\ref{tab:rl-hparams}.

\begin{table}[h]
\centering
\footnotesize
\setlength{\tabcolsep}{3pt}
\begin{tabular}{ll}
\toprule
\multicolumn{2}{l}{\textit{Shared by all RL methods}} \\
\midrule
Learning rate & $5\times10^{-6}$ \\
LR schedule & cosine, $3\%$ warmup \\
KL coefficient $\beta_{\mathrm{KL}}$ & $0.02$ \\
Rollouts per prompt & $8$ \\
Prompts per step & $4$ ($32$ completions) \\
Sampling temperature & $1.0$ \\
Completion budget & $1{,}024$ tokens \\
LoRA & as in SFT ($r{=}64$, $\alpha{=}128$) \\
Steps, Private-Science & $500$ \\
Steps, SAS & $200$ \\
\midrule
\multicolumn{2}{l}{\textit{\methodB{} only:} $\beta_\text{drift}$ (marks) / $\gamma_\Phi$} \\
\midrule
Private-Biology & $0.75$ / $0.3$ \\
Private-Physics & $0.5$ / $0.15$ \\
SAS-History & $0.75$ / $0.05$ \\
SAS-Geography & $0.75$ / $0.10$ \\
SAS-Physics & $0.75$ / $0.10$ \\
\bottomrule
\end{tabular}
\caption{RL hyperparameters. The upper block is shared by GRPO, DGPO, InT-RL, and \methodB{}. The lower block gives the per-dataset belief-shaping parameters of \methodB{}.}
\label{tab:rl-hparams}
\end{table}

\paragraph{Compute budget}
All RL runs use a single GPU (H200 141\,GB) with the rollout engine co-located. SFT uses two GPUs. We compare the two phases separately on Private-Biology, the largest dataset.
\emph{Phase 1 (data construction).} \methodA{} and InT start from the same pool of $20$ sampled rollouts per training response and the same $55{,}278$ incorrect rollouts. InT spends $\approx 8$ GPU-hours proposing interventions and sampling guided continuations. \methodA{} spends $\approx 9$ GPU-hours: $\approx 2.5$ for posterior-belief probing, $\approx 4$ for candidate selection and the grounding audit, and $\approx 2.5$ for sampling and validating atomic revisions. The probe reads the mark posterior at every step boundary of a rollout ($5.2$ boundaries on average). We obtain all of them in a single forward pass by appending one answer scaffold per boundary and using a block attention mask with matching position indices, so that each scaffold attends only to its own prefix. The resulting SFT pools hold $56{,}595$ samples for \methodA{} ($33{,}958$ golden rollouts and $22{,}637$ revisions) and $29{,}822$ for InT. A retained rollout can contribute more than one revised candidate step, so the number of revisions slightly exceeds the number of repaired rollouts implied by the retention rate of Appendix~\ref{app:counterfactual}. The two pools contain comparable numbers of generated revisions, and the difference is the golden anchor set, which is drawn from the shared rollout pool at no generation cost. SFT cost is proportional to pool size and identical per sample.
\emph{Phase 2 (RL).} \methodB{} and GRPO from the same \methodA{} checkpoint differ only by the frozen-probe read on each rollout group, which adds about $2$\,s to a $12$\,s optimiser step: $\approx 1.9$ versus $1.7$ GPU-hours for the $500$-step run. Base-initialised GRPO and DGPO incur only this RL cost. On the smaller datasets every phase takes at most a few GPU-hours for either method.

\section{Candidate Step Selection Procedure}
\label{app:selection}

Algorithm~\ref{alg:selection} formalises the two-stage selection rule of \S\ref{sec:method-phase-a}. The belief-shift criterion alone forms the shortlist. The grounding audit then re-ranks within that shortlist, promoting weakly grounded steps, and never excludes a step. In particular, a step with a large negative belief shift and strong grounding remains fully eligible.

\begin{algorithm}[h]
\caption{Candidate step selection in \methodA}
\label{alg:selection}
\begin{algorithmic}[1]
\Require incorrect rollout $s_{1:T}$ ($m \neq m^*$); belief shifts $\{\Delta\epsilon_j\}$; grounding scores $\{G_j^{\rho}, G_j^{a}, \widetilde{G}_j^{p}\}$; shortlist size $k$; candidate budget $n$; near-tie margin $\tau$; dataset-wide 25th percentiles $p_{25}(\cdot)$
\State $\mathcal{C} \gets \{\, j : \Delta\epsilon_j < -\tau \,\}$ \Comment{belief-shift criterion alone}
\State $\mathcal{C} \gets$ top-$k$ of $\mathcal{C}$ by most-negative $\Delta\epsilon_j$ \Comment{shortlist}
\For{$j \in \mathcal{C}$}
  \State $w_j \gets \big[G_j^{\rho}\!<\!p_{25}(G^{\rho})\big] \wedge \big[G_j^{a}\!<\!p_{25}(G^{a})\big] \wedge \big[\widetilde{G}_j^{p}\!<\!p_{25}(\widetilde{G}^{p})\big]$
\EndFor
\State stably reorder $\mathcal{C}$ so steps with $w_j{=}1$ precede those with $w_j{=}0$; ties keep the $\Delta\epsilon$ order \Comment{re-rank only; no step removed}
\State \Return first $\min(n, |\mathcal{C}|)$ steps of $\mathcal{C}$
\end{algorithmic}
\end{algorithm}

\section{Validity of the Frozen-Probe Reference}
\label{app:probe-kl}

\methodB reads the per-step mark belief $\hat{m}_k$ from a \emph{frozen} copy of the \methodA policy and penalises its drift from gold. We freeze the reference for two reasons. (i)~\textbf{Non-gameability:} RL could lower the shaping penalty by shifting its own belief \emph{readout} rather than by grading more faithfully. A fixed, exogenous reference removes this shortcut. (ii)~\textbf{Calibration:} the \methodA policy is already a competent grader, so its step-boundary mark posterior is a meaningful target for ``where belief should sit,'' unlike the raw policy. This design is sound only if the trained policy's belief remains well-described by the frozen readout, so we verify this directly.

We probe the mark posterior $\hat p_k(m)$ (Eq.~\ref{eq:probe}) at every step boundary of $200$ Private-Biology test rollouts under the frozen \methodA and under each SFT-initialised RL policy, and measure their divergence (Table~\ref{tab:probe-kl}). The \methodB policy stays very
close to the frozen reference (mean $\mathrm{KL}{=}0.026$, $\mathrm{JSD}{=}0.0055$) and is flat across chain position, confirming that the frozen probe remains a faithful proxy for the policy's belief. Plain GRPO, the same SFT initialisation and KL-to-init term, but \emph{without} belief shaping, drifts about twice as far ($\mathrm{KL}{=}0.064$), and increasingly so deeper in the chain. Belief shaping thus anchors the \emph{intermediate} trajectory to the calibrated reference, not merely the final mark.

\begin{table}[h]
\centering
\small
\setlength{\tabcolsep}{4.5pt}
\begin{tabular}{lccccc}
\toprule
& & & \multicolumn{3}{c}{JSD by position} \\
\cmidrule(lr){4-6}
Policy & KL & JSD & early & mid & late \\
\midrule
\methodB & \textbf{0.026} & \textbf{0.0055} & 0.0049 & 0.0055 & 0.0061 \\
GRPO      & 0.064          & 0.0119          & 0.0099 & 0.0133 & 0.0132 \\
\bottomrule
\end{tabular}
\caption{Step-boundary belief divergence between each SFT-initialised RL policy and the \emph{frozen} \methodA probe reference. KL and JSD are means over boundaries. The last three columns give JSD averaged within early/mid/late chain terciles. \methodB stays $\sim$2$\times$ closer to the reference and is flat across position, whereas GRPO drifts further and grows deeper in the chain.}
\label{tab:probe-kl}
\end{table}

\section{Masked-Support Audit Validation}
\label{app:support-validation}

We validate the masked-support audit against external step labels produced by a stronger annotator model (not human annotations), which tag each grading step as a \emph{grounding} step (a pointwise judgement read off the answer against the rubric) or a \emph{synthesis} step (one that composes prior conclusions). These model-generated labels validate the grounding signal itself. The causal question of whether selected steps are error-causing is addressed separately in Appendix~\ref{app:counterfactual}. \autoref{fig:support-validation} shows two clean dissociations. Grounding steps occupy the high Answer-Grounding region (panel~a), whereas synthesis steps collapse to $G^a\approx0$; conversely, synthesis steps carry markedly higher Prefix Grounding (panel~b, mode at $G^p\approx1.5$ vs.\ a near-zero mode for grounding steps). Both dissociations are highly significant (\autoref{tab:support-significance}): Answer Grounding separates the two classes with a large effect ($r{=}0.71$, $p<10^{-300}$) and Prefix Grounding likewise ($r{=}0.58$, $p<10^{-300}$), and both survive collapsing to per-rollout means. Rubric Grounding, in contrast, is statistically significant but practically negligible ($r{=}0.08$): both step types lean equally on the rubric, so the discriminating axes are \emph{which evidence} a step reads (the answer, for grounding steps) and \emph{whether} it integrates prior reasoning (the prefix, for synthesis steps). This confirms that $G^a$ and  $G^p$ behave as intended and are well separated from zero.

\begin{table}[h]
\centering
\small
\setlength{\tabcolsep}{4.5pt}
\begin{tabular}{lccrc}
\toprule
Signal & Grounding & Synthesis & \multicolumn{1}{c}{$r$} & $p$ \\
     & (median)  & (median)  & \multicolumn{1}{c}{}    &     \\
\midrule
 $G^a$ & \textbf{0.72} & 0.08          & 0.71 & $<\!10^{-300}$ \\
 $G^\rho$ & 0.89          & 0.83          & 0.08 & $5.6\!\times\!10^{-17}$ \\
 $G^p$ & 0.66          & \textbf{1.64} & 0.58 & $<\!10^{-300}$ \\
\bottomrule
\end{tabular}
\caption{Significance of the masked-grounding dissociations
(Figure~\ref{fig:support-validation}). $p$ is a one-sided Mann--Whitney $U$ test in the hypothesised direction (Grounding $>$ Synthesis for  $G^a$,  $G^\rho$; Synthesis $>$ Grounding for PG);
$r$ is the rank-biserial effect size.  $G^a$ and  $G^p$ dissociate with large effects. $G^\rho$ is significant but negligible ($r{=}0.08$). All three remain significant when collapsed to per-rollout means ( $G^a$,  $G^p$ $p\!\approx\!0$; $G^\rho$ $p{=}5.8\!\times\!10^{-36}$), ruling out within-rollout dependence as the driver.}
\label{tab:support-significance}
\end{table}

\begin{figure*}[h]
\centering
\includegraphics[width=0.85\textwidth]{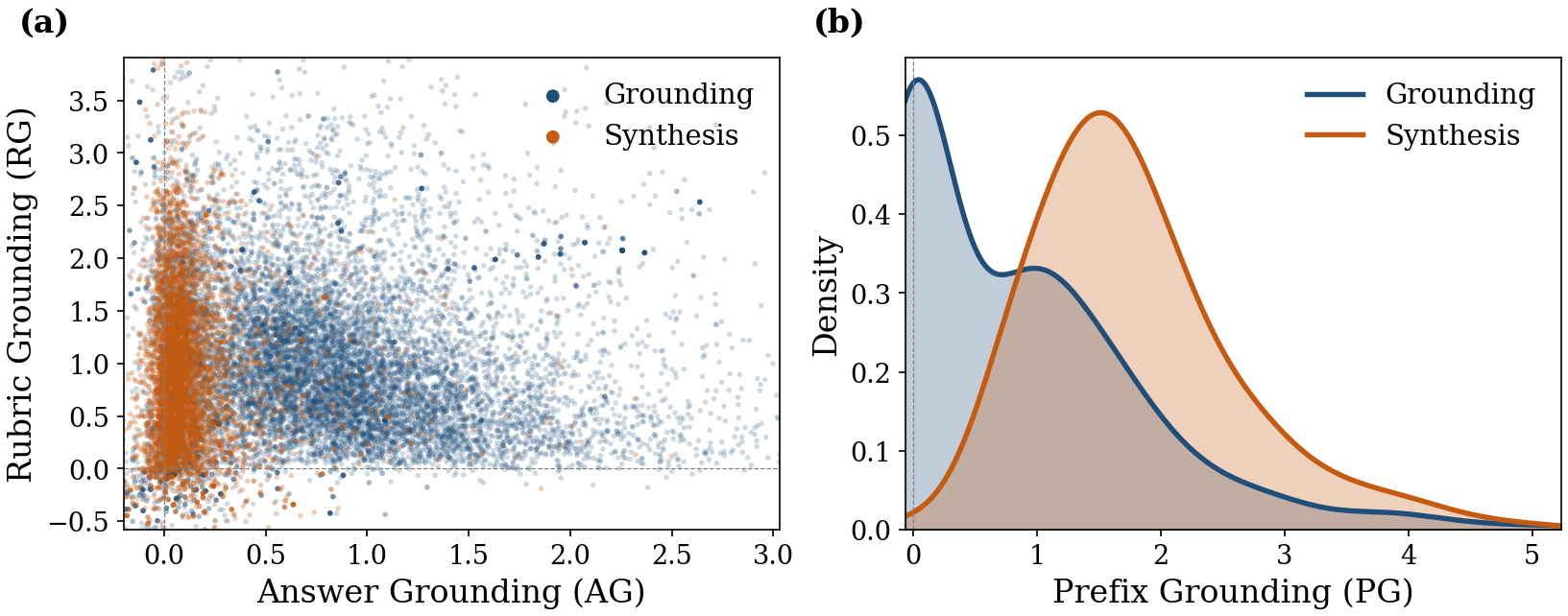}
\caption{Grounding audit distribution with different external step label (grounding, synthesis).}
\label{fig:support-validation}
\end{figure*}

\section{Counterfactual Validation of Step Selection}
\label{app:counterfactual}

We test whether the steps selected by \method{} are actually error-causing, in the counterfactual sense that repairing the selected step recovers the correct final mark. For each incorrect rollout, we compare three selection rules under an otherwise identical pipeline. \textbf{Random} draws a random step matched to \method{}'s relative-position distribution. \textbf{Belief shift only} applies the shortlist criterion without grounding re-ranking. The full \method{} rule applies both. Every arm uses the same rewriter, the same $N{=}4$ sampled continuations, and the same gold-mark filter, so differences are attributable to the choice of step alone. \textbf{Success@4} is the fraction of rewrites in which at least one continuation reaches the gold mark. \textbf{Retention} is the fraction of incorrect rollouts that yield a usable SFT target.

\begin{table}[h]
\centering
\small
\setlength{\tabcolsep}{5pt}
\begin{tabular}{llcc}
\toprule
\textbf{Dataset} & \textbf{Selection rule} & \textbf{S@4} & \textbf{Ret.} \\
\midrule
\multirow{3}{*}{Private-Biology} & Random & 30.0 & 33.1 \\
 & Belief shift only & 32.7 & 36.6 \\
 & \method & \textbf{33.5} & \textbf{37.2} \\
\midrule
\multirow{3}{*}{SAS-History} & Random & 16.6 & 18.5 \\
 & Belief shift only & 17.1 & 19.6 \\
 & \method & \textbf{22.0} & \textbf{23.6} \\
\bottomrule
\end{tabular}
\caption{Counterfactual repair comparison of step-selection rules. S@4 = Success@4 (\%). Ret.\ = Retention (\%).}
\label{tab:counterfactual}
\end{table}

As shown in Table~\ref{tab:counterfactual}, \method{}-selected steps recover the gold mark more often than position-matched random steps on both datasets. On SAS-History, \method{} raises the repair rate from $16.6\%$ to $22.0\%$, a relative gain of $1.33\times$, and yields $1.28\times$ more usable SFT examples. Private-Biology moves in the same direction with a smaller margin. The advantage also survives training: removing the localiser costs $4.3$ QWK points on SAS-History (Table~\ref{tab:ablation-sft}).

Outcome filtering is nonetheless costly: $62.8\%$ of incorrect rollouts on Private-Biology and $76.4\%$ on SAS-History yield no corrected example. This costs sampling compute rather than data validity, because every retained trajectory is verified against the gold mark. The resulting selection bias toward repairable cases is discussed in the Limitations section.

\section{Question-Level Significance Protocol}
\label{app:significance}

The pooled analysis of Table~\ref{tab:sas-sig} treats the question as the unit of independence. All $37$ held-out SAS questions are pooled with equal weight, so the pooled QWK differs from the subject-macro averages of Table~\ref{tab:main-sas}, which first average within subjects. Confidence intervals are obtained with a cluster bootstrap that resamples entire question clusters, never splitting responses to the same question across resamples. Comparisons against \texttt{Base} are paired: each method's per-question scores are differenced against \texttt{Base} on the same questions, so the tests condition on between-question variance even where marginal intervals overlap. Holm correction is applied across the six comparisons against \texttt{Base}. The pooled analysis complements the per-subject averages of Table~\ref{tab:main-sas}.

\section{Subject-Level Error Analysis}
\label{app:variance}

The smallest SAS subjects contain only $25$--$30$ test responses, so we draw no statistical conclusions from per-subject splits. The remaining variance is systematic. Across all $73$ prediction disagreements between \methodA{} and \texttt{InT-SFT}, $66\%$ are final score-band or credit-synthesis errors rather than missed evidence or rubric mapping, and the subjects stress exactly that mechanism because they differ sharply in supervision density and score range. In particular, $44\%$ of SAS-Physics evaluation responses carry a gold score of zero, so the grader must repeatedly withhold all credit.

\begin{table}[h]
\centering
\small
\setlength{\tabcolsep}{6pt}
\begin{tabular}{lccc}
\toprule
\textbf{System} & \textbf{History} & \textbf{Geography} & \textbf{Physics} \\
\midrule
\texttt{Base}    & 67 & 43 & 55 \\
\texttt{InT-SFT} & \textbf{78} & 43 & 0 \\
\methodA         & 72 & \textbf{57} & \textbf{64} \\
\bottomrule
\end{tabular}
\caption{Percentage of zero-score SAS responses that each system correctly assigns zero.}
\label{tab:zero-score}
\end{table}

Table~\ref{tab:zero-score} reports the percentage of zero-score responses each system correctly assigns zero. \texttt{InT-SFT} never withholds full credit on Physics, and its Physics QWK falls to $0.427$, while \methodA{} retains $0.701$ (Table~\ref{tab:main-sas}). This follows from the training targets: \texttt{InT} inserts a corrective redirection, which biases the continuation toward finding credit, whereas \method{} rewrites the erroneous rubric judgement itself, so the model also learns when credit must be withheld.

\section{Rule-Faithfulness Intervention}

\subsection{Intervention Strategies}
\label{app:int-strategies}
We probe rule-faithfulness using six deterministic, gold-computable edits to the mark scheme, grouped into two families that correspond to the two Private-Biology scheme types. For each edit we recompute the reference (gold) mark under the edited scheme, regrade the \emph{same} student response with the edited scheme supplied in-prompt, and compare the model's score change to the gold change via the Rule-Sensitivity Ratio $\mathrm{RSR}=\overline{|\Delta\mathrm{pred}|}/\overline{|\Delta\mathrm{gold}|}$
(\S\ref{sec:exp-intervention}).

\paragraph{Level-of-response (LoR) band shifts.} For LoR questions, the scheme maps each response-quality level to a mark band. We apply: \textbf{A1 (uniform $+2$)}: adding a constant $2$ marks to every level ($\overline{|\Delta\mathrm{gold}|}{=}1.85$); \textbf{A2 (expand $\times2$)}: doubling each level's mark, stretching inter-level spacing ($\overline{|\Delta\mathrm{gold}|}{=}3.00$); \textbf{B1 (differential)}: a non-uniform per-level shift (e.g. $+1/+2/+2$) from the lowest to the highest band.

\paragraph{Points-total (PTS) rescales.} For PTS questions, the scheme awards a fixed mark per creditable point. We multiply every point's value by a constant factor: \textbf{$\times0.5$} ($\overline{|\Delta\mathrm{gold}|}{=}0.35$), \textbf{$\times1.5$} ($0.90$), and \textbf{$\times2.0$} ($1.25$).

\paragraph{Floor confound.} A prediction of $0$ is a fixed point of every edit
(it cannot move), so a grader can appear faithful merely by predicting $0$. We
therefore decompose RSR into a \emph{floor rate} (fraction of perturbed
predictions equal to $0$) and an \emph{in-rubric RSR} (RSR computed only over
rows with a non-zero perturbed prediction), and report both.

\subsection{Full Intervention Results}
\label{app:int_full_results}

\begin{table*}[t]
\centering
\scriptsize
\setlength{\tabcolsep}{7pt}
\begin{tabular}{l cc cc cc cc cc cc cc}
\toprule
& \multicolumn{2}{c}{Base} & \multicolumn{2}{c}{GRPO} & \multicolumn{2}{c}{DGPO}
& \multicolumn{2}{c}{InT-SFT} & \multicolumn{2}{c}{InT-RL}
& \multicolumn{2}{c}{\methodA} & \multicolumn{2}{c}{\methodB} \\
\cmidrule(lr){2-3}\cmidrule(lr){4-5}\cmidrule(lr){6-7}\cmidrule(lr){8-9}\cmidrule(lr){10-11}\cmidrule(lr){12-13}\cmidrule(lr){14-15}
\textbf{Rule edit} & FR & R$_{ir}$ & FR & R$_{ir}$ & FR & R$_{ir}$ & FR & R$_{ir}$ & FR & R$_{ir}$ & FR & R$_{ir}$ & FR & R$_{ir}$ \\
\midrule
\multicolumn{15}{l}{\textit{LoR}}\\
LoR-A1 ($+2$)      &  8.9 & 1.00 &  5.8 & 0.81 &  6.1 & 0.83 & 10.2 & 0.82 &  9.7 & 0.78 &  6.5 & 0.87 &  6.4 & 0.86 \\
LoR-A2 ($\times2$) & 12.3 & 1.13 &  8.9 & 0.96 &  8.8 & 0.96 & 10.6 & 0.99 & 10.3 & 0.92 &  7.8 & 0.96 &  6.9 & 0.94 \\
LoR-B1 (diff)      &  9.0 & 1.14 &  6.8 & 0.95 &  6.7 & 0.94 & 10.9 & 0.91 &  9.1 & 0.87 &  6.7 & 0.96 &  6.5 & 0.95 \\
\midrule
\multicolumn{15}{l}{\textit{PTS}}\\
PTS ($\times0.5$)  & 14.2 & 1.47 & 13.3 & 1.35 & 13.5 & 1.35 & 31.9 & 1.06 & 29.6 & 1.06 & 16.8 & 1.14 & 17.2 & 1.11 \\
PTS ($\times1.5$)  & 25.6 & 0.16 & 21.4 & 0.13 & 21.7 & 0.13 & 26.0 & 0.27 & 25.6 & 0.29 & 21.9 & 0.35 & 20.7 & 0.33 \\
PTS ($\times2.0$)  & 27.5 & 0.11 & 21.7 & 0.09 & 22.8 & 0.13 & 27.8 & 0.20 & 25.2 & 0.25 & 21.0 & 0.29 & 21.7 & 0.25 \\
\bottomrule
\end{tabular}
\caption{Full per-strategy rule-faithfulness on Private-Biology. Each model column reports \textbf{FR} (floor rate, \% predicted $0$) and \textbf{R}$_{ir}$ (in-rubric RSR; $1.0$=faithful).}
\label{tab:int-full-bio}
\end{table*}

Table \ref{tab:int-full-bio} gives the full breakdown. \textbf{LoR edits: all graders are near-faithful.} In-rubric RSR stays in $[0.78,1.14]$ and floor rates are low ($6$--$12\%$); the model tracks band shifts as intended. The \method family is the most faithful trained family here (in-rubric RSR $0.86$--$0.96$, closest to $1$) and floors the least ($6.4$--$7.8\%$), whereas \texttt{Base} mildly over-reacts ($1.13$--$1.14$ on A2/B1). \textbf{PTS rescales: unfaithful for everyone.} Multiplicative point rescaling collapses in-rubric RSR to $0.1$--$0.35$ on the up-rescales ($\times1.5,\times2.0$) and inflates floor rates to $21$--$32\%$, evidence of absolute-count anchoring that no method removes. \texttt{InT} floors hardest ($26$--$32\%$); \methodA retains the highest PTS in-rubric RSR among the trained arms ($0.25$--$0.35$ on the up-rescales). The decomposition also exposes a confound in the headline RSR: \texttt{InT}'s competitive aggregate RSR on PTS is partly a flooring artefact (its high floor rate removes responsive rows), which the floor-rate / in-rubric split makes explicit. Signed averages can likewise flatter \texttt{Base}, whose over-reactions on LoR and under-reactions on PTS partly cancel. The AE-RSR aggregation in Table~\ref{tab:int-decomp} removes this cancellation.

\section{Prompt}
\label{sec:appendix_prompt}

We document the core prompt behind \method{}'s Phase-1 atomic rewriting under locality constraint (shown in Figure~\ref{fig:prompt_rewrite}). For space we show only its structural skeleton: the section headers, the per-instance privileged inputs (rendered as \texttt{\{$\cdot$\}} placeholders), and the output contract; the verbose per-section guidance is elided (\texttt{[\dots]}). 

\begingroup
\setlength{\textfloatsep}{6pt plus 2pt minus 2pt}
\setlength{\floatsep}{6pt plus 2pt minus 2pt}
\setlength{\intextsep}{6pt plus 2pt minus 2pt}
\setlength{\dbltextfloatsep}{6pt plus 2pt minus 2pt}
\setlength{\dblfloatsep}{6pt plus 2pt minus 2pt}
\setlength{\abovecaptionskip}{2pt}
\setlength{\belowcaptionskip}{0pt}

\makeatletter
\setlength{\@dblfptop}{0pt}
\setlength{\@dblfpsep}{6pt}
\setlength{\@dblfpbot}{0pt}
\setlength{\@fptop}{0pt}
\setlength{\@fpsep}{6pt}
\setlength{\@fpbot}{0pt}
\makeatother

\begin{figure*}[t]
\centering
\begin{promptbox}
SYSTEM: Refine a wrong grading attempt by fixing exactly ONE substep.
        The rest of the chain regenerates after the rewrite.

# Grading task
  {PROBLEM}

# Oracle reference
  correct mark = {GOLD} / {TOTAL}
  [oracle only; never state the mark digit or its spelled-out word]

# The full grading attempt
  {NUMBERED_ATTEMPT}

# Target for rewriting: Substep {i}
  "{TARGET_STEP_TEXT}"

# Diagnostic hints for Substep {i}
  {DIAGNOSTIC_HINT_BULLETS}
  [from the internal-state locator (C1); a checklist to CONSIDER, not satisfy]

# Per-criterion rubric analysis (REFERENCE)
  {RCI_ANALYSIS}
  [privileged input (C2); informs reasoning, never copied into the output]

# Atomicity + locality
  [serve the SAME local role as the target step; emit multiple atomic
   substeps only when the prefix has already synthesised criteria]

# Style
  [one claim per step; match the chain's voice; no "(1)(2)" markers,
   no covered/partial verdict labels, no enumeration]

# Output format
  <rewrite location="{i}" span="1|2|3">
  step 1: <first step text>
  step 2: <optional>
  </rewrite>
\end{promptbox}
\caption{\method{} Phase-1 \emph{Rule-Aware Atomic Rewriter} prompt (structural skeleton; per-section guidance elided as \texttt{[\dots]}). Braced tokens \texttt{\{$\cdot$\}} are filled per instance. The diagnostic hints come from the internal-state locator (\textbf{C1}) and the per-criterion rubric analysis is the privileged input (\textbf{C2}); neither the gold mark nor the rubric block's structure may appear in the rewrite.}
\label{fig:prompt_rewrite}
\end{figure*}

\endgroup

\end{document}